\documentclass[11pt,letterpaper]{article}
\pdfoutput=1
\usepackage[hyperref]{udst2018}
\usepackage{times}
\usepackage{latexsym}
\usepackage{amsmath}
\usepackage{url}
\usepackage{amssymb}
\usepackage{amsfonts}
\usepackage{graphicx}
\usepackage{tabularx}
\usepackage{multirow}

\aclfinalcopy 
 
\title{An improved neural network model for joint POS tagging and dependency parsing}

\author{Dat Quoc Nguyen \and Karin Verspoor  \\
School of Computing and Information Systems \\
The University of Melbourne, Australia \\
{\tt \{dqnguyen, karin.verspoor\}@unimelb.edu.au}
}

\date{}

\begin{document}

\maketitle

\begin{abstract}
We propose a novel neural network model for joint part-of-speech (POS) tagging and dependency parsing.  Our model  extends the well-known BIST graph-based dependency parser \citep{TACL885} by incorporating a BiLSTM-based tagging component to produce automatically predicted POS tags for the parser. 
On the benchmark English Penn treebank, our model obtains strong UAS and LAS scores at 94.51\% and 92.87\%, respectively, producing 1.5+\%  absolute improvements to the  BIST graph-based  parser, and also obtaining a state-of-the-art POS tagging accuracy at 97.97\%. Furthermore, 
experimental results on parsing 61 ``big'' Universal Dependencies treebanks  from raw texts show that our model outperforms the baseline UDPipe \citep{udpipe:2017}  with 0.8\% higher average POS tagging score and 3.6\% higher average LAS  score. 
In addition, with our model,  we also obtain state-of-the-art downstream task scores for biomedical event extraction and opinion analysis applications.   

Our code is available together with all pre-trained models  at: \url{https://github.com/datquocnguyen/jPTDP}.


\end{abstract}

\begin{figure*}[!t]
\centering
\includegraphics[width=14cm]{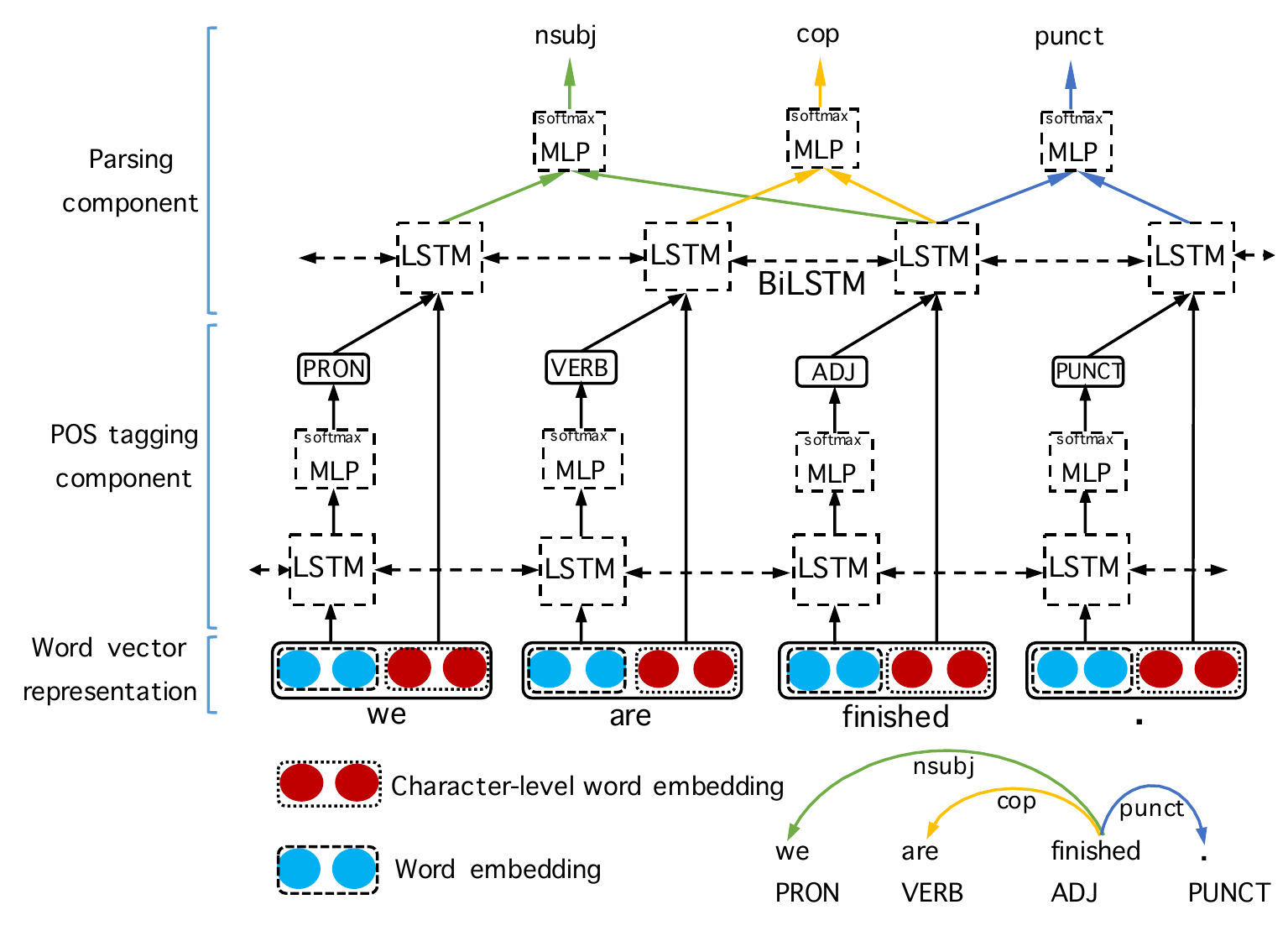} 
\vspace{-10pt}
\caption{ Illustration of our new model for joint POS tagging and graph-based dependency parsing.}
\label{fig:Architecture}
\end{figure*}

\section{Introduction}

Dependency parsing -- a key research topic in natural language processing (NLP) in the last decade \cite{Buchholz2006,Nivre07,Kubler2009} -- has  also been demonstrated to be extremely useful in many applications such as relation extraction \citep{culotta-sorensen:2004:ACL,bunescu-mooney:2005:HLTEMNLP}, semantic parsing \citep{Q16-1010} and machine translation \citep{galley-manning:2009:ACLIJCNLP}. 
In general, dependency parsing models can be categorized as graph-based  
\cite{McDonald2005OLT} and transition-based  
\cite{Yamada2003,Nivre2003}.  
 Most traditional graph- or transition-based models define a set of core and combined features  \citep{E06-1011,Nivre2007,Bohnet2010,zhang-nivre:2011:ACL-HLT2011},  while recent state-of-the-art  models propose neural network architectures to handle feature-engineering
\citep{dyer-EtAl:2015:ACL-IJCNLP,TACL885,DozatM17,ma-hovy:2017:I17-1}. 
 
 Most traditional and neural network-based parsing models use automatically predicted POS  tags as essential features. However, POS taggers are not perfect,  resulting in error propagation problems. Some work has attempted to avoid using POS tags for dependency parsing \citep{dyer-EtAl:2015:ACL-IJCNLP,ballesterosEMNLP,K17-3022}, however,   to achieve the strongest parsing scores these methods still require automatically assigned POS tags. Alternatively, joint POS tagging and dependency parsing has also attracted a lot of  attention in  NLP community as it could help improve  both tagging and parsing results over independent modeling  \citep{Li:2011:JMC:2145432.2145557,hatori-EtAl:2011:IJCNLP-2011,Lee:2011:DMJ:2002472.2002584,bohnet-nivre:2012:EMNLP-CoNLL,zhang-EtAl:2015:NAACL-HLT1,zhang-weiss:2016:P16-1,YangZLSYF18}.

In this paper, we present a novel  neural network-based model for jointly learning POS tagging and dependency paring. Our joint model extends the well-known BIST graph-based dependency parser \citep{TACL885}  with an additional lower-level BiLSTM-based tagging component. In particular, this tagging component generates predicted POS tags for the parser component.  Evaluated on the benchmark English Penn treebank test Section 23, our model produces a 1.5+\%  absolute improvement over the  BIST graph-based  parser with a  strong UAS score of 94.51\%  and LAS score of 92.87\%; and also obtaining a state-of-the-art POS tagging accuracy of 97.97\%. In addition,  multilingual parsing  experiments from raw texts on   61 ``big'' Universal Dependencies treebanks   \citep{udst:overview} 
 show that our model outperforms the baseline UDPipe \citep{udpipe:2017}   with 0.8\% higher average POS tagging score, 3.1\% higher UAS and  3.6\% higher LAS. Furthermore, experimental results on downstream task applications \citep{epe18} show that our joint model helps produce state-of-the-art scores for biomedical event extraction and opinion analysis. 

\section{Our joint model}

This section presents our  model for joint POS tagging and graph-based dependency parsing. Figure \ref{fig:Architecture} illustrates the architecture of our joint model which can be viewed as a two-component mixture of a tagging component and a parsing component. Given word tokens in an input sentence, the tagging component uses a BiLSTM to learn ``latent'' feature vectors representing these word tokens. Then the tagging component feeds these feature vectors into a multilayer perceptron with one hidden layer (MLP)  to predict POS tags.  The parsing component then uses another BiLSTM to learn another set of latent feature representations, based on both the input word tokens and the predicted POS tags.  These  latent feature representations are   fed into a MLP to decode  dependency arcs and another MLP to label the predicted dependency arcs.

\subsection{Word vector representation}\label{ssec:wv}
Given an input sentence $s$ consisting of $n$ word tokens $w_1, w_2, ..., w_n$, we represent each $i^{th}$ word  $w_i$ in $s$ by a vector $\mathbf{e}_{i}$. We obtain $\mathbf{e}_{i}$ by concatenating word embedding $\mathbf{e}^{(\textsc{w})}_{w_i}$ and character-level word embedding $\mathbf{e}^{(\textsc{c})}_{w_i}$:

\begin{equation}
\mathbf{e}_{i} =  \mathbf{e}^{(\textsc{w})}_{w_i} \circ \mathbf{e}^{(\textsc{c})}_{w_i} 
\end{equation}

\noindent Here, each word type $w$ in the training data is represented by a real-valued word embedding $\mathbf{e}^{(\textsc{w})}_{w}$. 
Given  the word type $w$ consisting of $k$ characters $w = c_1c_2...c_k$ where each j$^{th}$ character in $w$  is represented by a character embedding  $\mathbf{c}_{j}$, 
we use a sequence BiLSTM ($\mathrm{BiLSTM}_{\text{seq}}$) to learn its character-level  vector representation \citep{ballesterosEMNLP,plankP16}.  The input to $\mathrm{BiLSTM}_{\text{seq}}$ is the sequence of  $k$ character embeddings $\mathbf{c}_{1:k}$, 
and the output is a concatenation of outputs of a forward LSTM (LSTM\textsubscript{f}) reading the input in its regular order and a reverse LSTM (LSTM\textsubscript{r}) reading the input in reverse:

\medskip
\noindent\centerline{\small 
$\mathbf{e}^{(\textsc{c})}_{w}  = \text{BiLSTM\textsubscript{seq}}(\mathbf{c}_{1:k}) \\ = \text{LSTM\textsubscript{f}}(\mathbf{c}_{1:k}) \circ \text{LSTM\textsubscript{r}}(\mathbf{c}_{k:1})$
}

\subsection{Tagging component}

We feed the sequence of vectors $\mathbf{e}_{1:n}$ with an additional context position  index $i$   into another BiLSTM ($\mathrm{BiLSTM}_{\text{pos}}$), resulting in latent feature vectors $\boldsymbol{v}_{i}^{(\text{pos})}$ each representing the $i^{\text{th}}$ word $w_i$ in  $s$:

\begin{equation}
\boldsymbol{v}_{i}^{(\text{pos})} =   \mathrm{BiLSTM}_{\text{pos}} (\mathbf{e}_{1:n}, i)
\end{equation}

We use a MLP  with softmax output ($\mathrm{MLP}_{\text{pos}}$) on top of the $\mathrm{BiLSTM}_{\text{pos}}$ to predict POS tag of each word in $s$. The number of nodes in the output layer of this $\mathrm{MLP}_{\text{pos}}$ is the number of POS tags. Given  $\boldsymbol{v}_{i}^{(\text{pos})}$, we compute an output vector as:

  \begin{equation}
{\boldsymbol{\vartheta}}_i = \mathrm{MLP}_{\mathrm{pos}}(\boldsymbol{v}_{i}^{(\text{pos})}) 
\end{equation}

Based on output vectors ${\boldsymbol{\vartheta}}_i$, we then compute the cross-entropy objective loss \textbf{$\mathcal{L}_{\text{POS}}(\hat{\mathbf{t}}, \mathbf{t})$}, in which $\hat{\mathbf{t}}$ and $\mathbf{t}$ are the sequence of predicted POS tags and sequence of gold POS tags of words in the input sentence $s$, respectively \citep{Goldberg16}. Our tagging component thus can be viewed as a simplified version of the  POS tagging model proposed by \citet{plankP16}, without  their additional auxiliary loss for rare words.

\subsection{Parsing component}
Assume that $p_1, p_2, ..., p_n$ are the predicted POS tags produced by the tagging component for the input words.
We represent each $i^{th}$ predicted POS tag by a vector embedding $\mathbf{e}^{(\textsc{p})}_{p_i}$. 
We then create a sequence of vectors $\boldsymbol{x}_{1:n}$ in which each $\boldsymbol{x}_{i}$ is produced by concatenating the  POS tag embedding $\mathbf{e}^{(\textsc{p})}_{p_i}$ and the word vector representation  $\mathbf{e}_{i}$:

\begin{equation}
\mathbf{x}_{i} = \mathbf{e}^{(\textsc{p})}_{p_i}  \circ   \mathbf{e}_{i}=  \mathbf{e}^{(\textsc{p})}_{p_i}  \circ \mathbf{e}^{(\textsc{w})}_{w_i} \circ \mathbf{e}^{(\textsc{c})}_{w_i} 
\end{equation}

We feed the sequence of vectors $\mathbf{x}_{1:n}$ with an additional index $i$ into a  BiLSTM ($\mathrm{BiLSTM}_{\text{dep}}$), resulting in latent feature vectors $\boldsymbol{v}_{i}$ as follows:

\begin{equation}
\boldsymbol{v}_{i} =   \mathrm{BiLSTM}_{\text{dep}} (\mathbf{x}_{1:n}, i)
\label{equal:v_i}
\end{equation}

Based on latent feature vectors $\boldsymbol{v}_{i}$, we follow a common arc-factored parsing approach to decode dependency arcs \citep{McDonald2005OLT}. 
 In particular,  a dependency tree  can be formalized as a directed graph.  An arc-factored parsing approach learns the scores of the arcs in the graph \citep{Kubler2009}. 
Here, we score an arc by using a MLP with a one-node output layer ($\mathrm{MLP}_{\mathrm{arc}}$) on top of the $\mathrm{BiLSTM}_{\text{dep}}$:

\vspace{-15pt}
  \begin{flalign}
&\mathrm{score}_{\mathrm{arc}}(i, j)  \\
&= \mathrm{MLP}_{\mathrm{arc}}\big( \boldsymbol{v}_{i} \circ \boldsymbol{v}_{j} \circ (\boldsymbol{v}_{i} \ast \boldsymbol{v}_{j}) \circ |\boldsymbol{v}_{i} -  \boldsymbol{v}_{j}| \big) \nonumber
\end{flalign}

\noindent where $(\boldsymbol{v}_{i} \ast \boldsymbol{v}_{j})$ and $|\boldsymbol{v}_{i} -  \boldsymbol{v}_{j}|$ denote the element-wise product and the absolute element-wise difference, respectively;  and $\boldsymbol{v}_{i}$ and $\boldsymbol{v}_{j}$ are correspondingly  the  latent feature vectors associating to the $i^{th}$ and $j^{th}$ words in $s$,  computed by Equation \ref{equal:v_i}.

Given the arc scores, we use the \citet{Eisner:1996}'s decoding  algorithm to find the highest scoring projective parse tree:

\vspace{-10pt}
\begin{equation}
\mathrm{score}(s) =   \underset{\hat{y} \in \mathcal{Y}(s)}{\mathrm{argmax}} \sum\limits_{(h, m) \in \hat{y}} \mathrm{score}_{\mathrm{arc}}(h, m)
\end{equation}

\noindent where $\mathcal{Y}(s)$ is the set of all possible dependency trees for the input sentence $s$ while $\mathrm{score}_{\mathrm{arc}}(h, m)$ measures the score of the arc between the head $h^{\text{th}}$ word and the modifier $m^{\text{th}}$ word in  $s$.

Following  \citet{TACL885}, we compute a  margin-based hinge loss \textbf{$\mathcal{L}_{\text{ARC}}$} with loss-augmented inference  to maximize the margin between the gold unlabeled parse tree and the highest scoring incorrect tree.

For predicting dependency relation type of a head-modifier arc, we use another MLP with softmax output ($\mathrm{MLP}_{\mathrm{rel}}$) on top of the $\mathrm{BiLSTM}_{\text{dep}}$. Here, the number of the nodes in the output layer of this $\mathrm{MLP}_{\mathrm{rel}}$  is the number of dependency relation types.  Given an arc $(h, m)$, we compute an  output vector as:

\vspace{-15pt}
  \begin{flalign}
&{\textbf{v}}_{(h,m)}  \\
&= \mathrm{MLP}_{\mathrm{rel}}\big( \boldsymbol{v}_{h} \circ \boldsymbol{v}_{m} \circ (\boldsymbol{v}_{h} \ast \boldsymbol{v}_{m}) \circ |\boldsymbol{v}_{h} -  \boldsymbol{v}_{m}| \big) \nonumber
\end{flalign}

Based on output vectors ${\textbf{v}}_{(h,m)}$, we also compute another cross-entropy objective loss \textbf{$\mathcal{L}_{\text{REL}}$}  for relation type prediction, using only the gold labeled parse tree.    

Our parsing component  can be viewed as an extension of the  BIST graph-based dependency model  \citep{TACL885}, where we additionally incorporate  the character-level vector representations of words.

\subsection{Joint model training}

The  training objective loss of our joint model is  the sum of the
POS tagging loss $\mathcal{L}_{\text{POS}}$, the structure loss $\mathcal{L}_{\text{ARC}}$ and
the  relation labeling loss $\mathcal{L}_{\text{REL}}$: 

\begin{equation}
\mathcal{L} = \mathcal{L}_{\text{POS}} + \mathcal{L}_{\text{ARC}}  + \mathcal{L}_{\text{REL}}
\end{equation}

\noindent The model parameters, including word embeddings, character
embeddings, POS  embeddings, three one-hidden-layer MLPs and three BiLSTMs, are
learned  to minimize the sum $\mathcal{L}$  of the losses.

Most neural network-based joint models for  POS tagging and dependency parsing are transition-based approaches \citep{alberti-EtAl:2015:EMNLP,zhang-weiss:2016:P16-1,YangZLSYF18}, while our model is a graph-based method. In addition, the joint model JMT \citep{hashimoto-EtAl:2017:EMNLP2017} defines its dependency parsing task as a head selection task which produces a probability distribution over possible heads for each word \citep{zhang-cheng-lapata:2017:EACLlong}. 

Our   model is the successor of the joint model jPTDP v1.0 \citep{nguyen-dras-johnson:2017:K17-3} which is also  a graph-based method.  However, unlike our  model, jPTDP v1.0 uses a BiLSTM to learn ``shared'' latent feature vectors which are then used for both  POS tagging and dependency parsing tasks, 
rather than using two separate layers.
As mentioned in Section \ref{sec:ud}, our  model generally outperforms jPTDP v1.0 with 2.5+\% LAS  improvements on universal dependencies (UD) treebanks.

\subsection{Implementation details}\label{ssec:implement}

Our model  is released as jPTDP v2.0, available  at \url{https://github.com/datquocnguyen/jPTDP}. Our jPTDP v2.0 is implemented using  \textsc{DyNet} v2.0 \citep{dynet} with a fixed random seed.\footnote{\url{https://github.com/clab/dynet}}  Word embeddings are initialized either randomly or by pre-trained word vectors, while character and POS tag embeddings are randomly initialized. For learning character-level word embeddings, we use one-layer $\mathrm{BiLSTM}_{\text{seq}}$, and  set the size of LSTM hidden states  to be equal to the vector size of character embeddings.  

We  apply dropout \citep{JMLR:v15:srivastava14a} with a 67\% keep probability to the inputs of  BiLSTMs and MLPs. Following \citet{iyyer-EtAl:2015:ACL-IJCNLP} and \citet{TACL885},  we also apply \textit{word dropout} to learn  an embedding for unknown words: we replace each word token $w$ appearing $\#(w)$ times in the training set  with a special ``unk'' symbol with probability $\mathsf{p}_{unk}(w) = \frac{0.25}{0.25 + \#(w)}$. This procedure only involves the word embedding part in the input word vector representation.

We optimize the objective loss using Adam  \citep{KingmaB14} with an initial learning rate at 0.001  and no mini-batches.  For training, we run for 30 epochs, and restart the Adam optimizer and anneal its initial learning rate at a proportion of 0.5 every 10 epochs.  
We  evaluate the \textit{mixed accuracy} of correctly assigning POS tag together with dependency arc and relation type on the development set after each training epoch. We choose the model with the highest mixed accuracy on the development set, which is then applied to the test set for the evaluation phase.

For all experiments presented in this paper, we use 100-dimensional word embeddings, 50-dimensional character embeddings and 100-dimensional POS tag embeddings. We also fix the number of hidden nodes in MLPs at 100. Due to limited computational resource, for experiments presented in Section \ref{sec:enptb}, we perform a minimal grid search of hyper-parameters to select the number of $\mathrm{BiLSTM}_{\text{pos}}$ and $\mathrm{BiLSTM}_{\text{dep}}$ layers from $\{1, 2\}$ 
  and the size of LSTM hidden states  in each layer from $\{128, 256\}$. For experiments presented in sections \ref{sec:ud} and \ref{sec:epe}, we fix the number of BiLSTM layers at 2 and the size of  hidden states  at 128.

\section{Experiments on English Penn treebank}\label{sec:enptb}

\paragraph{Experimental setup:} We evaluate our model using the  English  WSJ Penn treebank \cite{Marcus93building}. We follow a standard data split to use   sections 02-21 for training, Section 22 for development and Section 23 for test \citep{chen-manning:2014:EMNLP2014}, employing the Stanford  conversion toolkit v3.3.0 to generate dependency trees  with  Stanford basic dependencies \citep{deMarneffe:2008:STD:1608858.1608859}.  

Word embeddings are initialized by 100-dimensional GloVe word vectors  pre-trained on  Wikipedia and Gigaword \citep{pennington-socher-manning:2014:EMNLP2014}.\footnote{\scriptsize\url{https://nlp.stanford.edu/projects/glove}} As mentioned in Section \ref{ssec:implement}, we perform a minimal grid search of hyper-parameters and find that 
the highest mixed accuracy on the development set is obtained when using 2 BiLSTM layers and 256-dimensional LSTM hidden states (in Table \ref{tab:devscores}, we  present   scores  obtained on  the development set when using 2 BiLSTM layers). 

\begin{table}[!t]
\centering
\setlength{\tabcolsep}{0.25em}
\begin{tabular}{c|lll|ll}
\hline
\multirow{2}{*}{\bf \#states} &  \multicolumn{3}{c|}{\bf With punctuations} & \multicolumn{2}{c}{\bf Without pun.}  \\
\cline{2-6}
 & POS & UAS & LAS & UAS & LAS \\
\hline
 128 & \textbf{97.64} & 93.68 & 92.11 & 94.42 & 92.61 \\
 256 & 97.63 &  \textbf{93.89} & \textbf{92.33} & \textbf{94.63} & \textbf{92.82}\\
 \hline
 \hline
 \multicolumn{4}{r|}{ \protect{\citet{chen-manning:2014:EMNLP2014}} } & 92.0 & 89.7 \\
 \multicolumn{4}{r|}{ \protect{\citet{dyer-EtAl:2015:ACL-IJCNLP}} } & 93.2 & 90.9 \\
  \multicolumn{4}{r|}{BIST-graph [K\&G16]} & 93.3 & 91.0 \\
 \multicolumn{4}{r|}{ \citet{zhang-cheng-lapata:2017:EACLlong}} & 94.30 & 91.95 \\
  \multicolumn{4}{r|}{\citet{ma-hovy:2017:I17-1}} & 94.77 & 92.66\\
  \multicolumn{4}{r|}{ \protect{\citet{DozatM17}} } & \textbf{95.24} & \textbf{93.37} \\
 \hline
\end{tabular}
\caption{Results on the development set. \textbf{\#states} and ``\textbf{Without pun.}'' denote the size of LSTM hidden states and the scores computed without punctuations, respectively. ``POS'' indicates the POS tagging accuracy. [K\&G16]   denotes results reported in  \protect{\citet{TACL885}}.
}
\label{tab:devscores}
\end{table}

\begin{table}[!t]
\centering
\resizebox{7.5cm}{!}{
\setlength{\tabcolsep}{0.3em}
\begin{tabular}{l|lll}
\hline
{\bf Model} & \textbf{POS} & \textbf{UAS} & \textbf{LAS} \\
\hline 
 \protect{\citet{chen-manning:2014:EMNLP2014}} & 97.3 & 91.8 & 89.6 \\
 \protect{\citet{dyer-EtAl:2015:ACL-IJCNLP}} & 97.3   & 93.1 & 90.9 \\
 \citet{weiss-EtAl:2015:ACL-IJCNLP}  & 97.44 & 93.99 & 92.05 \\
  BIST-graph [K\&G16] & 97.3  & 93.1 & 91.0  \\
  BIST-transition [K\&G16] & 97.3 & 93.9 & 91.9   \\
  \citet{kuncoro-EtAl:2016:EMNLP2016} & 97.3  &94.26 & 92.06 \\
  \citet{andor-EtAl:2016:P16-1} & 97.44  & 94.61 & 92.79 \\ 
 \citet{zhang-cheng-lapata:2017:EACLlong} & 97.3  & 94.10 & 91.90 \\
 \citet{ma-hovy:2017:I17-1} & 97.3  & 94.88 & 92.98\\
 \protect{\citet{DozatM17}} & 97.3  &  \textbf{95.44} & \textbf{93.76} \\
 \protect{\citet{DozatM17}} [$\bullet$] & 97.3  &  {95.66} & {94.03} \\
  
 \hline
 \hline
 \citet{bohnet-nivre:2012:EMNLP-CoNLL} [$\star$]& 97.42 & 93.67 & 92.68  \\
 \citet{alberti-EtAl:2015:EMNLP} & 97.44 & 94.23 & 92.36 \\
 \citet{zhang-weiss:2016:P16-1} & \_ & 93.43 & 91.41\\
 \citet{hashimoto-EtAl:2017:EMNLP2017}& \_  & \textbf{94.67} & \textbf{92.90} \\
  \citet{YangZLSYF18}  & 97.54 & 94.18 & 92.26\\
\hline
Our   model  & \textbf{97.97}& 94.51 & 92.87 \\
\hline
\end{tabular}
}
\caption{Results on the test set. POS tagging accuracies are computed on all tokens. UAS and LAS are computed without punctuations.    [$\bullet$]: the treebank was converted with the Stanford  conversion toolkit v3.5.0. 
[$\star$]:  the treebank was converted with the head rules of \citet{Yamada2003}. For both [$\bullet$] and [$\star$], 
 obtained parsing scores are just for reference, not for  comparison.}
\label{tab:wsjsults}
\end{table}

\paragraph{Main results:}  Table  \ref{tab:wsjsults} compares 
 our UAS and LAS scores  on the test set with previous published results in terms of the dependency annotations.\footnote{\citet{choe-charniak:2016:EMNLP2016}  reported  the highest UAS score at 95.9\% and LAS score at 94.1\%  to date on the test set, using the Stanford conversion toolkit v3.3.0 to convert the output constituent trees into  dependency representations.}    
 The first 11 rows present scores of dependency parsers in which POS tags were predicted by using an  external POS tagger such as the Stanford tagger \citep{N03-1033}. 
 The last 6 rows present scores for joint models. 
Clearly, our model produces very competitive parsing results.  
 In particular, our model obtains a  UAS score at 94.51\% and a LAS score at 92.87\% which are about  1.4\% and 1.9\% absolute higher than UAS and LAS scores   of the  BIST graph-based model \citep{TACL885}, respectively. Our model also does better than the previous transition-based joint models in  \citet{alberti-EtAl:2015:EMNLP},  \citet{zhang-weiss:2016:P16-1} and \citet{YangZLSYF18}, while obtaining similar UAS and LAS scores to the joint  model JMT proposed by \citet{hashimoto-EtAl:2017:EMNLP2017}. 
 
 We achieve 0.9\% lower parsing scores than the state-of-the-art  dependency parser of \citet{DozatM17}.
While also a BiLSTM- and graph-based model, it  uses a more sophisticated  attention mechanism  ``\textit{biaffine}'' for better decoding  dependency arcs and  relation types.
 In future work, we will extend our model with the  \textit{biaffine} attention  mechanism to investigate the benefit for our model. Other differences are that they use a higher dimensional representation than ours, but rely on predicted POS tags.

We also obtain a state-of-the-art POS tagging accuracy at 97.97\%  on the test Section 23, which is about 0.4+\% higher than those by \citet{bohnet-nivre:2012:EMNLP-CoNLL}, \citet{alberti-EtAl:2015:EMNLP} and \citet{YangZLSYF18}. Other previous joint models did not mention their specific POS tagging accuracies.\footnote{\citet{hashimoto-EtAl:2017:EMNLP2017} showed that JMT obtains a POS tagging accuracy of 97.55\% on WSJ sections 22-24.}

\section{UniMelb in the CoNLL 2018 shared task on UD parsing}\label{sec:ud}

Our  UniMelb team participated with  jPTDP v2.0 in the CoNLL 2018 shared task   on parsing 82 treebank test sets (in 57 languages) from raw text to universal dependencies \citep{udst:overview}. 
The 82 treebanks are taken from  UD  v2.2 \citep{11234/1-2837}, where 61/82 test sets are for  ``big'' UD treebanks  for which both training and development data sets are available and 5/82 test sets are extra ``parallel''  test sets in languages where another big treebank exists. In addition, 7/82 test sets are for  ``small'' UD treebanks for which development data is not available. The remaining 9/82 sets  are in low-resource languages without training data or with a few gold-annotation sample sentences. 

For the 7 small treebanks without development data available, we  split training data into two parts with a ratio 9:1, and then use the larger part for training and the smaller part for development.  For each big or small treebank, we train a joint model for \textit{universal} POS tagging and dependency parsing, using a fixed random seed and a fixed set of hyper-parameters as mentioned in Section \ref{ssec:implement}.\footnote{We initialize word embeddings by 100-dimensional pre-trained  vectors from \citet{11234/1-1989}. For a language where pre-trained word vectors are not available in \citet{11234/1-1989}, word embeddings  are randomly initialized.} 
We evaluate the mixed accuracy on the development set after each training epoch, and select the model with the highest mixed accuracy.

For parsing from raw text to universal dependencies, we employ  CoNLL-U test files pre-processed by the baseline UDPipe 1.2 \citep{udpipe:2017}. Here, 
we utilize the tokenization, word and sentence segmentation  predicted by  UDPipe 1.2.   For 68 big and small treebank test files, we use the corresponding  trained joint models. We use the joint models trained for \textit{cs\_pdt}, \textit{en\_ewt}, \textit{fi\_tdt}, \textit{ja\_gsd} and \textit{sv\_talbanken} to process 5 parallel test files  \textit{cs\_pud}, \textit{en\_pud}, \textit{fi\_pud}, \textit{ja\_modern} and \textit{sv\_pud}, respectively. Since we do not focus on low-resource languages, we employ  the baseline UDPipe 1.2 to process 9 low-resource treebank test files. The final test runs are carried out on the TIRA  platform \citep{tira}.

\begin{table}[!t]
\centering
\resizebox{7.5cm}{!}{
\setlength{\tabcolsep}{0.3em}
\begin{tabular}{ll|lllll}
\hline
\multicolumn{2}{c|}{\multirow{2}{*}{\bf System}}& \textbf{All} & \textbf{Big} & \textbf{PUD} & \textbf{Small} & \textbf{Low}    \\
& & (82) & (61) & (5) & (7) & (9) \\
\hline
\multirow{3}{*}{\rotatebox[origin=c]{90}{UPOS}}  
&  UDPipe 1.2 & 87.32 & 93.71 & 85.23 & \textbf{87.36} & 45.20 \\
  & UniMelb  & \textbf{87.90} & \textbf{94.50} & \textbf{85.33} & 87.12 & 45.20 \\
  \cline{2-7} 
  \cline{2-7} 
  &  goldseg. & \_ & 95.63 & 90.21 & 87.64 & \_ \\
\hline
\multirow{3}{*}{\rotatebox[origin=c]{90}{UAS}}  
&  UDPipe 1.2 & 71.64 & 78.78 & 71.22 & 63.17 & 30.08\\
  & UniMelb  &  \textbf{74.16} & \textbf{81.83} & \textbf{73.17} & \textbf{64.71} & 30.08\\
    \cline{2-7} 
  \cline{2-7} 
  & goldseg.  & \_ & 85.01 & 81.81 & 67.46  & \_  \\
\hline
\multirow{3}{*}{\rotatebox[origin=c]{90}{LAS}}  
&  UDPipe 1.2 & 65.80 & 74.14 & 66.63 & 55.01 & 17.17\\
  & UniMelb  & \textbf{68.65} & \textbf{77.69} & \textbf{68.72} & \textbf{56.12} & 17.17\\
      \cline{2-7} 
  \cline{2-7} 
 & goldseg. & \_ & 80.68 &  75.03 & 58.65 & \_ \\
\hline
\end{tabular}
}
\caption{ Official macro-average F1 scores computed on all tokens for  UniMelb and  the baseline UDPipe 1.2 in the CoNLL 2018 shared task on UD parsing from raw texts \citep{udst:overview}. ``UPOS'' denotes the universal POS tagging score. ``All'', ``Big'', ``PUD'', ``Small'' and ``Low''  refer 
to the macro-average scores over all 81, 61 big treebank, 5 parallel, 7 small treebank and 9 low-resource treebank test sets, respectively.  
``goldseg.'' denotes the scores of our jPTDP v2.0 model regarding gold segmentation, detailed in Table \ref{tab:Scoreswithgoldseg}.}
\label{tab:conll18results}
\end{table}

\begin{table*}[!ht]
\centering
\resizebox{15.75cm}{!}{
\setlength{\tabcolsep}{0.2em}
\begin{tabular}{l|l|ccc||l|l|ccc}
\hline
\textbf{Treebank} & \textbf{Code} & \textbf{UPOS} & \textbf{UAS} & \textbf{LAS} & \textbf{Treebank} & \textbf{Code} & \textbf{UPOS} & \textbf{UAS} & \textbf{LAS} \\
\hline
Afrikaans-AfriBooms & af\_afribooms & 95.73 & 82.57 & 78.89 & Italian-ISDT & it\_isdt & 98.01 & 92.33 & 90.20 \\
Ancient\_Greek-PROIEL & grc\_proiel & 96.05 & 77.57 & 72.84 & Italian-PoSTWITA & it\_postwita & 95.41 & 84.20 & 79.11 \\
Ancient\_Greek-Perseus & grc\_perseus & 88.95 & 65.09 & 58.35 & Japanese-GSD & ja\_gsd & 97.27 & 94.21 & 92.02 \\
Arabic-PADT & ar\_padt & 96.33 & 86.08 & 80.97 & Japanese-Modern \textbf{[p]} & ja\_modern & 70.53 & 66.88 & 49.51 \\
Basque-BDT & eu\_bdt & 93.62 & 79.86 & 75.07 & Korean-GSD & ko\_gsd & 93.35 & 81.32 & 76.58 \\
Bulgarian-BTB & bg\_btb & 98.07 & 91.47 & 87.69 & Korean-Kaist & ko\_kaist & 93.53 & 83.59 & 80.74 \\
Catalan-AnCora & ca\_ancora & 98.46 & 90.78 & 88.40 & Latin-ITTB & la\_ittb & 98.12 & 82.99 & 79.96 \\
Chinese-GSD & zh\_gsd & 93.26 & 82.50 & 77.51 & Latin-PROIEL & la\_proiel & 95.54 & 74.95 & 69.76 \\
Croatian-SET & hr\_set & 97.42 & 88.74 & 83.62 & Latin-Perseus \textbf{[s]}& la\_perseus & 82.36 & 57.21 & 46.28 \\
Czech-CAC & cs\_cac & 98.87 & 89.85 & 87.13 & Latvian-LVTB & lv\_lvtb & 93.53 & 81.06 & 76.13 \\
Czech-FicTree & cs\_fictree & 97.98 & 88.94 & 85.64 & North\_Sami-Giella \textbf{[s]} & sme\_giella & 87.48 & 65.79 & 58.09 \\
Czech-PDT & cs\_pdt & 98.74 & 89.64 & 87.04 & Norwegian-Bokmaal & no\_bokmaal & 97.73 & 89.83 & 87.57 \\
Czech-PUD \textbf{[p]} & cs\_pud & 96.71 & 87.62 & 82.28 & Norwegian-Nynorsk & no\_nynorsk & 97.33 & 89.73 & 87.29 \\
Danish-DDT & da\_ddt & 96.18 & 82.17 & 78.88 & Norwegian-NynorskLIA \textbf{[s]} & no\_nynorsklia & 85.22 & 64.14 & 54.31 \\
Dutch-Alpino & nl\_alpino & 95.62 & 86.34 & 82.37 & Old\_Church\_Slavonic-PROIEL & cu\_proiel & 93.69 & 80.59 & 73.93 \\
Dutch-LassySmall & nl\_lassysmall & 95.21 & 86.46 & 82.14 & Old\_French-SRCMF & fro\_srcmf & 95.12 & 86.65 & 81.15 \\
English-EWT & en\_ewt & 95.48 & 87.55 & 84.71 & Persian-Seraji & fa\_seraji & 96.66 & 88.07 & 84.07 \\
English-GUM & en\_gum & 94.10 & 84.88 & 80.45 & Polish-LFG & pl\_lfg & 98.22 & 95.29 & 93.10 \\
English-LinES & en\_lines & 95.55 & 80.34 & 75.40 & Polish-SZ & pl\_sz & 97.05 & 90.98 & 87.66 \\
English-PUD \textbf{[p]} & en\_pud & 95.25 & 87.49 & 84.25 & Portuguese-Bosque & pt\_bosque & 96.76 & 88.67 & 85.71 \\
Estonian-EDT & et\_edt & 96.87 & 85.45 & 82.13 & Romanian-RRT & ro\_rrt & 97.43 & 88.74 & 83.54 \\
Finnish-FTB & fi\_ftb & 94.53 & 86.10 & 82.45 & Russian-SynTagRus & ru\_syntagrus & 98.51 & 91.00 & 88.91 \\
Finnish-PUD \textbf{[p]} & fi\_pud & 96.44 & 87.54 & 84.60 & Russian-Taiga \textbf{[s]} & ru\_taiga & 85.49 & 65.52 & 56.33 \\
Finnish-TDT & fi\_tdt & 96.12 & 86.07 & 82.92 & Serbian-SET & sr\_set & 97.40 & 89.32 & 85.03 \\
French-GSD & fr\_gsd & 97.11 & 89.45 & 86.43 & Slovak-SNK & sk\_snk & 95.18 & 85.88 & 81.89 \\
French-Sequoia & fr\_sequoia & 97.92 & 89.71 & 87.43 & Slovenian-SSJ & sl\_ssj & 97.79 & 88.26 & 86.10 \\
French-Spoken & fr\_spoken & 94.25 & 79.80 & 73.45 & Slovenian-SST & sl\_sst \textbf{[s]} & 89.50 & 66.14 & 58.13 \\
Galician-CTG & gl\_ctg & 97.12 & 85.09 & 81.93 & Spanish-AnCora & es\_ancora & 98.57 & 90.30 & 87.98 \\
Galician-TreeGal \textbf{[s]} & gl\_treegal & 93.66 & 77.71 & 71.63 & Swedish-LinES & sv\_lines & 95.51 & 83.60 & 78.97 \\
German-GSD & de\_gsd & 94.07 & 81.45 & 76.68 & Swedish-PUD \textbf{[p]} & sv\_pud & 92.10 & 79.53 & 74.53 \\
Gothic-PROIEL & got\_proiel & 93.45 & 79.80 & 71.85 & Swedish-Talbanken & sv\_talbanken & 96.55 & 86.53 & 83.01 \\
Greek-GDT & el\_gdt & 96.59 & 87.52 & 84.64 & Turkish-IMST & tr\_imst & 92.93 & 70.53 & 62.55 \\
Hebrew-HTB & he\_htb & 96.24 & 87.65 & 82.64 & Ukrainian-IU & uk\_iu & 95.24 & 83.47 & 79.38 \\
Hindi-HDTB & hi\_hdtb & 96.94 & 93.25 & 89.83 & Urdu-UDTB & ur\_udtb & 93.35 & 86.74 & 80.44 \\
Hungarian-Szeged & hu\_szeged & 92.07 & 76.18 & 69.75 & Uyghur-UDT & ug\_udt & 87.63 & 76.14 & 63.37 \\
Indonesian-GSD & id\_gsd & 93.29 & 84.64 & 77.71 & Vietnamese-VTB & vi\_vtb & 87.63 & 67.72 & 58.27 \\
\cline{6-10}
Irish-IDT \textbf{[s]} & ga\_idt & 89.74 & 75.72 & 65.78 &\multicolumn{2}{r|}{\textbf{Average}}& 94.49 &  83.11 &  78.18\\
\hline
\end{tabular}
}
\caption{UPOS, UAS and LAS scores computed on all tokens of our jPTDP v2.0 model
regarding gold-standard segmentation on 73 CoNLL-2018 shared task test sets ``Big'', ``PUD'' and ``Small'' -- UD v2.2 \citep{11234/1-2837}.  \textbf{[p]} and \textbf{[s]} denote the ``PUD'' extra parallel and small test sets, respectively. For each treebank, a joint model is trained using a fixed set of hyper-parameters as mentioned in Section \ref{ssec:implement}.}
\label{tab:Scoreswithgoldseg}
\end{table*}

Table \ref{tab:conll18results}  presents our   results
in the CoNLL 2018 shared task on multilingual parsing from raw texts to universal dependencies \citep{udst:overview}.  Over all 82 test sets, we outperform the baseline UDPipe 1.2 with 0.6\% absolute higher average UPOS F1 score and 2.5+\%  higher average UAS and LAS F1 scores. In particular, for the ``big'' category consisting of 61 treebank test sets, we obtain 0.8\% higher UPOS  and 3.1\% higher UAS and 3.6\% higher LAS  than UDPipe 1.2.

Our (UniMelb) official LAS-based rank is at $14^{th}$ place while the baseline  UDPipe 1.2 is at  $18^{th}$ place over total 26 participating systems.\footnote{\url{http://universaldependencies.org/conll18/results.html}} However, 
 it is difficult  to make a clear comparison between our jPTDP v2.0  and the parsing models used in other top  systems. 
 Several better participating systems simply reuse the state-of-the-art \textit{biaffine} dependency parser  \citep{DozatM17,dozat-qi-manning:2017:K17-3},  constructing
ensemble models or developing  treebank concatenation strategies to obtain larger training data, which is likely to produce better scores than ours  \citep{udst:overview}. 
 
 Recall that the shared task focuses on parsing from raw texts.  
 Most higher-ranking systems aim to improve the pre-processing steps of tokenization\footnote{\url{http://universaldependencies.org/conll18/results-tokens.html}}, word\footnote{\url{http://universaldependencies.org/conll18/results-words.html}} and sentence\footnote{\url{http://universaldependencies.org/conll18/results-sentences.html}} segmentation, resulting in significant improvements in final parsing scores. For example, in the CoNLL 2017 shared task on UD parsing \citep{K17-3001}, UDPipe 1.2 obtained 0.1+\% higher average tokenization and word segmentation scores and 0.2\% higher average sentence segmentation score than UDPipe 1.1, resulting in 1+\% improvement in the final  average LAS F1 score while both UDPipe 1.2 and UDPipe 1.1  shared exactly the same remaining components.  
 Utilizing  better pre-processors, as used in other participating systems,  should likewise improve our final parsing scores. 
 
In Table \ref{tab:conll18results}, we also present our average UPOS, UAS and LAS accuracies with respect to (w.r.t.) gold-standard tokenization, word and sentence segmentation. 
For more details and future comparison,   Table \ref{tab:Scoreswithgoldseg}  presents the UPOS, UAS and LAS scores w.r.t.\  gold-standard   segmentation, obtained by  jPTDP v2.0 on each UD v2.2--CoNLL 2018 shared task test
set. Compared to the scores presented in Table 3 in \citet{nguyen-dras-johnson:2017:K17-3}   on overlapped treebanks, our model jPTDP v2.0 generally produces 2.5+\% improvements in UAS and LAS scores to  jPTDP v1.0 \citep{nguyen-dras-johnson:2017:K17-3}.

\section{UniMelb in the  EPE 2018 campaign}\label{sec:epe}

Our  UniMelb team also participated with jPTDP v2.0 in the 2018 Extrinsic Parser Evaluation (EPE) campaign \citep{epe18}.\footnote{\url{http://epe.nlpl.eu}}  The EPE 2018 campaign runs in collaboration with the CoNLL 2018 shared task, which
aims to evaluate dependency parsers by comparing their performance on three downstream tasks: biomedical  event extraction \cite{Bjo:Gin:Sal:17}, negation resolution \cite{Lap:Oep:Ovr:17} and opinion analysis \cite{Johansson:17}. 
Here, participants only need to provide parsing outputs of English raw texts used in these downstream tasks; the campaign organizers then  compute
end-to-end downstream task scores. 
General background can be also found in the first EPE edition 2017 \citep{epe17}. 

\begin{table*}[t]
\centering

\begin{tabular}{l|lll|l||lll|l}
\hline
\multirow{2}{*}{\bf Task}  & \multicolumn{4}{c||}{\bf Development set} & \multicolumn{4}{c}{\bf Evaluation set}\\ 
\cline{2-9}
& Pre. & Rec. & F1 & SP17 & Pre. & Rec. & F1 & SP17 \\
\hline
Event extraction & 57.87	& 51.20	& \textbf{54.33}\textsubscript{1} & 52.67\textsubscript{54.59} & 58.52	& 49.43	& \textbf{53.59}\textsubscript{1} & 50.29\textsubscript{50.23} \\
Negation resolution & 100 &	44.51 &	61.60\textsubscript{3} & \textbf{64.85}\textsubscript{65.37} & 100	& 41.83	& 58.99\textsubscript{3} & \textbf{65.13}\textsubscript{66.16} \\
Opinion analysis & 69.12	& 64.65	& \textbf{66.81}\textsubscript{1} & 66.63\textsubscript{68.53} & 66.67	& 62.88	& \textbf{64.72}\textsubscript{1} & 63.72\textsubscript{65.14} \\
\hline
Average & \_ & \_ & 60.91\textsubscript{1} & \textbf{61.38}\textsubscript{62.83} & \_ & \_ & 59.10\textsubscript{1} & \textbf{59.71}\textsubscript{60.51}\\
\hline
\end{tabular}
\caption{Downstream task scores Precision (Prec.), Recall (Rec.) and F1 for our UniMelb team. The \textit{subscript} in the F1 column denotes the unofficial rank of UniMelb over 17 participating teams at EPE 2018 \protect{\citep{epe18}}. ``SP17'' denotes the F1 scores obtained by the EPE 2017  system Stanford-Paris \protect{\cite{stanfordparis2017}} with respect to (w.r.t.) the Stanford basic dependencies. The \textit{subscript} in the SP17 column denotes the F1 scores obtained by Stanford-Paris w.r.t.\ the UD-v1-enhanced type of dependency representations, in which the average F1 score at 60.51 is the highest one at EPE 2017.}
\label{tab:epe18}
\end{table*}

Unlike  EPE 2017, the EPE 2018 campaign limited the training data to the English UD treebanks only. We unfortunately were unaware of this restriction during development of our model. Thus, we trained a jPTDP v2.0 model on dependency trees generated with the Stanford basic dependencies on a  combination of the WSJ treebank, sections 02-21, and the training split of the GENIA treebank \cite{I05-2038}. We used the fixed set of hyper-parameters as used for the CoNLL 2018 shared task as mentioned in Section \ref{ssec:implement}.\footnote{Word embeddings are initialized by the 100-dimensional pre-trained GloVe word vectors.} We then submitted the parsing outputs by running our trained model on the pre-processed tokenized and sentence-segmented data provided by the campaign on the TIRA platform.

Table \ref{tab:epe18} presents the  results  we  obtained for three downstream tasks at EPE 2018 \citep{epe18}. 
Since we employed external training data, our obtained scores are not officially  ranked.  In total 17 participating teams, we obtained the highest average F1 score  over the three downstream tasks (i.e., we ranked first, unofficially).  
In particular, we achieved the highest F1 scores for both biomedical event extraction and opinion analysis. 
 Our  results may be high because the training data we used is larger than the English UD treebanks used by other  teams.

Table \ref{tab:epe18} also presents scores from the Stanford-Paris team \cite{stanfordparis2017}---the first-ranked team at  EPE 2017  \citep{epe17}. 
Both EPE 2017 and 2018 campaigns use the same downstream task setups,  therefore the downstream task scores are directly comparable. 
Note that Stanford-Paris employed the state-of-the-art \textit{biaffine} dependency parser \cite{dozat-qi-manning:2017:K17-3} with larger training data. In particular, Stanford-Paris not only used the WSJ sections 02-21 and the training split of the GENIA treebank (as we did), but also included the Brown corpus. 
The downstream application of negation resolution  requires parsing of fiction, which is one the genres included in the Brown corpus. Hence it is reasonable that the Stanford-Paris team produced better negation resolution scores than we did. 

However, in terms of the Stanford basic dependencies, while we employ a less  accurate parsing model with smaller training data, we obtain higher downstream task scores for event extraction and opinion analysis than the Stanford-Paris team. Consequently, better intrinsic parsing performance does not always imply better extrinsic downstream application performance. Similar observations on the biomedical event extraction and opinion analysis tasks can also be found in \citet{NguyenV2018} and \citet{GomezRodriguez2017}, respectively. 
Further investigations of this pattern requires much deeper understanding of the architecture of the downstream task systems, which is left for future work.

\section{Conclusion}

In this paper, we have presented a novel neural network model for joint
POS tagging and graph-based dependency parsing. On the benchmark English WSJ Penn treebank, our model obtains strong parsing scores UAS at 94.51\% and LAS at 92.87\%, and a state-of-the-art POS tagging accuracy at 97.97\%. 

We also participated with our joint model in the CoNLL 2018 shared task on multilingual parsing from raw texts to universal dependencies, and obtained very competitive results. Specifically, using the same  CoNLL-U files pre-processed by UDPipe \citep{udpipe:2017},  our  model produced   0.8\% higher  POS tagging, 3.1\% higher UAS and  3.6\% higher LAS  scores  on average than UDPipe on 61 big UD treebank test sets. 
Furthermore, our model also helps obtain state-of-the-art downstream task scores for the biomedical event extraction and opinion analysis applications.  

We believe our joint model can serve as a new strong baseline for both intrinsic POS tagging and dependency parsing tasks as well as for extrinsic downstream applications. Our code and pre-trained models are available at: \url{https://github.com/datquocnguyen/jPTDP}.

\section*{Acknowledgments}   
This work was supported by the ARC Discovery Project DP150101550 and ARC Linkage Project LP160101469.

\bibliography{REFs}

\begin{thebibliography}{}
\expandafter\ifx\csname natexlab\endcsname\relax\def\natexlab#1{#1}\fi

\bibitem[{Alberti et~al.(2015)Alberti, Weiss, Coppola, and
  Petrov}]{alberti-EtAl:2015:EMNLP}
Chris Alberti, David Weiss, Greg Coppola, and Slav Petrov. 2015.
\newblock {Improved Transition-Based Parsing and Tagging with Neural Networks}.
\newblock In {\em Proceedings of EMNLP\/}. pages 1354--1359.

\bibitem[{Andor et~al.(2016)Andor, Alberti, Weiss, Severyn, Presta, Ganchev,
  Petrov, and Collins}]{andor-EtAl:2016:P16-1}
Daniel Andor, Chris Alberti, David Weiss, Aliaksei Severyn, Alessandro Presta,
  Kuzman Ganchev, Slav Petrov, and Michael Collins. 2016.
\newblock {Globally Normalized Transition-Based Neural Networks}.
\newblock In {\em Proceedings of ACL\/}. pages 2442--2452.

\bibitem[{Ballesteros et~al.(2015)Ballesteros, Dyer, and
  Smith}]{ballesterosEMNLP}
Miguel Ballesteros, Chris Dyer, and Noah~A. Smith. 2015.
\newblock {Improved Transition-based Parsing by Modeling Characters instead of
  Words with LSTMs}.
\newblock In {\em Proceedings of EMNLP\/}. pages 349--359.

\bibitem[{Bj{\"o}rne et~al.(2017)Bj{\"o}rne, Ginter, and
  Salakoski}]{Bjo:Gin:Sal:17}
Jari Bj{\"o}rne, Filip Ginter, and Tapio Salakoski. 2017.
\newblock {EPE}~2017: The {B}iomedical event extraction downstream application.
\newblock In {\em Proceedings of the EPE 2017 Shared Task\/}. pages 17--24.

\bibitem[{Bohnet(2010)}]{Bohnet2010}
Bernd Bohnet. 2010.
\newblock Very high accuracy and fast dependency parsing is not a
  contradiction.
\newblock In {\em Proceedings of COLING\/}. pages 89--97.

\bibitem[{Bohnet and Nivre(2012)}]{bohnet-nivre:2012:EMNLP-CoNLL}
Bernd Bohnet and Joakim Nivre. 2012.
\newblock {A Transition-Based System for Joint Part-of-Speech Tagging and
  Labeled Non-Projective Dependency Parsing}.
\newblock In {\em Proceedings of EMNLP-CoNLL\/}. pages 1455--1465.

\bibitem[{Buchholz and Marsi(2006)}]{Buchholz2006}
Sabine Buchholz and Erwin Marsi. 2006.
\newblock {CoNLL-X shared task on multilingual dependency parsing}.
\newblock In {\em Proceedings of CoNLL\/}. pages 149--164.

\bibitem[{Bunescu and Mooney(2005)}]{bunescu-mooney:2005:HLTEMNLP}
Razvan Bunescu and Raymond Mooney. 2005.
\newblock {A Shortest Path Dependency Kernel for Relation Extraction}.
\newblock In {\em Proceedings of HLT/EMNLP\/}. pages 724--731.

\bibitem[{Chen and Manning(2014)}]{chen-manning:2014:EMNLP2014}
Danqi Chen and Christopher Manning. 2014.
\newblock A fast and accurate dependency parser using neural networks.
\newblock In {\em Proceedings of EMNLP\/}. pages 740--750.

\bibitem[{Choe and Charniak(2016)}]{choe-charniak:2016:EMNLP2016}
Do~Kook Choe and Eugene Charniak. 2016.
\newblock {Parsing as Language Modeling}.
\newblock In {\em Proceedings of EMNLP\/}. pages 2331--2336.

\bibitem[{Culotta and Sorensen(2004)}]{culotta-sorensen:2004:ACL}
Aron Culotta and Jeffrey Sorensen. 2004.
\newblock {Dependency Tree Kernels for Relation Extraction}.
\newblock In {\em Proceedings of ACL\/}. pages 423--429.

\bibitem[{de~Lhoneux et~al.(2017)de~Lhoneux, Shao, Basirat, Kiperwasser,
  Stymne, Goldberg, and Nivre}]{K17-3022}
Miryam de~Lhoneux, Yan Shao, Ali Basirat, Eliyahu Kiperwasser, Sara Stymne,
  Yoav Goldberg, and Joakim Nivre. 2017.
\newblock {From Raw Text to Universal Dependencies - Look, No Tags!}
\newblock In {\em Proceedings of the CoNLL 2017 Shared Task\/}. pages 207--217.

\bibitem[{de~Marneffe and Manning(2008)}]{deMarneffe:2008:STD:1608858.1608859}
Marie-Catherine de~Marneffe and Christopher~D. Manning. 2008.
\newblock {The Stanford Typed Dependencies Representation}.
\newblock In {\em Proceedings of the Workshop on Cross-Framework and
  Cross-Domain Parser Evaluation\/}. pages 1--8.

\bibitem[{Dozat and Manning(2017)}]{DozatM17}
Timothy Dozat and Christopher~D. Manning. 2017.
\newblock {Deep Biaffine Attention for Neural Dependency Parsing}.
\newblock In {\em Proceedings of ICLR\/}.

\bibitem[{Dozat et~al.(2017)Dozat, Qi, and
  Manning}]{dozat-qi-manning:2017:K17-3}
Timothy Dozat, Peng Qi, and Christopher~D. Manning. 2017.
\newblock {Stanford's Graph-based Neural Dependency Parser at the CoNLL 2017
  Shared Task}.
\newblock In {\em Proceedings of the CoNLL 2017 Shared Task\/}. pages 20--30.

\bibitem[{Dyer et~al.(2015)Dyer, Ballesteros, Ling, Matthews, and
  Smith}]{dyer-EtAl:2015:ACL-IJCNLP}
Chris Dyer, Miguel Ballesteros, Wang Ling, Austin Matthews, and Noah~A. Smith.
  2015.
\newblock {Transition-Based Dependency Parsing with Stack Long Short-Term
  Memory}.
\newblock In {\em Proceedings of ACL-IJCNLP\/}. pages 334--343.

\bibitem[{Eisner(1996)}]{Eisner:1996}
Jason~M. Eisner. 1996.
\newblock {Three New Probabilistic Models for Dependency Parsing: An
  Exploration}.
\newblock In {\em Proceedings of COLING\/}. pages 340--345.

\bibitem[{Fares et~al.(2018)Fares, Oepen, Øvrelid, Bj{\"o}rne, and
  Johansson}]{epe18}
Murhaf Fares, Stephan Oepen, Lilja Øvrelid, Jari Bj{\"o}rne, and Richard
  Johansson. 2018.
\newblock The 2018 {S}hared {T}ask on {E}xtrinsic {P}arser {E}valuation. {O}n
  the downstream utility of {E}nglish universal dependency parsers.
\newblock In {\em {Proceedings of the CoNLL 2018 Shared Task}\/}. page to
  appear.

\bibitem[{Galley and Manning(2009)}]{galley-manning:2009:ACLIJCNLP}
Michel Galley and Christopher~D. Manning. 2009.
\newblock {Quadratic-Time Dependency Parsing for Machine Translation}.
\newblock In {\em Proceedings of ACL-IJCNLP\/}. pages 773--781.

\bibitem[{Ginter et~al.(2017)Ginter, Haji{\v c}, Luotolahti, Straka, and
  Zeman}]{11234/1-1989}
Filip Ginter, Jan Haji{\v c}, Juhani Luotolahti, Milan Straka, and Daniel
  Zeman. 2017.
\newblock \href{http://hdl.handle.net/11234/1-1989}{{{CoNLL} 2017 Shared Task -
  Automatically Annotated Raw Texts and Word Embeddings}}.
\newblock
  \href{http://hdl.handle.net/11234/1-1989}{http://hdl.handle.net/11234/1-1989}.

\bibitem[{Goldberg(2016)}]{Goldberg16}
Yoav Goldberg. 2016.
\newblock {A Primer on Neural Network Models for Natural Language Processing}.
\newblock {\em Journal of Artificial Intelligence Research\/} 57:345--420.

\bibitem[{G{\'o}mez-Rodr{\'i}guez et~al.(2017)G{\'o}mez-Rodr{\'i}guez,
  Alonso-Alonso, and Vilares}]{GomezRodriguez2017}
Carlos G{\'o}mez-Rodr{\'i}guez, Iago Alonso-Alonso, and David Vilares. 2017.
\newblock How important is syntactic parsing accuracy? an empirical evaluation
  on rule-based sentiment analysis.
\newblock {\em Artificial Intelligence Review\/} .

\bibitem[{Hashimoto et~al.(2017)Hashimoto, xiong, Tsuruoka, and
  Socher}]{hashimoto-EtAl:2017:EMNLP2017}
Kazuma Hashimoto, caiming xiong, Yoshimasa Tsuruoka, and Richard Socher. 2017.
\newblock {A Joint Many-Task Model: Growing a Neural Network for Multiple NLP
  Tasks}.
\newblock In {\em Proceedings of EMNLP\/}. pages 1923--1933.

\bibitem[{Hatori et~al.(2011)Hatori, Matsuzaki, Miyao, and
  Tsujii}]{hatori-EtAl:2011:IJCNLP-2011}
Jun Hatori, Takuya Matsuzaki, Yusuke Miyao, and Jun'ichi Tsujii. 2011.
\newblock {Incremental Joint POS Tagging and Dependency Parsing in Chinese}.
\newblock In {\em Proceedings of IJCNLP\/}. pages 1216--1224.

\bibitem[{Iyyer et~al.(2015)Iyyer, Manjunatha, Boyd-Graber, and
  Daum\'{e}~III}]{iyyer-EtAl:2015:ACL-IJCNLP}
Mohit Iyyer, Varun Manjunatha, Jordan Boyd-Graber, and Hal Daum\'{e}~III. 2015.
\newblock {Deep Unordered Composition Rivals Syntactic Methods for Text
  Classification}.
\newblock In {\em Proceedings of ACL-IJCNLP\/}. pages 1681--1691.

\bibitem[{Johansson(2017)}]{Johansson:17}
Richard Johansson. 2017.
\newblock {EPE}~2017: {T}he {T}rento--{G}othenburg opinion extraction system.
\newblock In {\em Proceedings of the EPE 2017 Shared Task\/}. pages 31--39.

\bibitem[{Kingma and Ba(2014)}]{KingmaB14}
Diederik~P. Kingma and Jimmy Ba. 2014.
\newblock {Adam: {A} Method for Stochastic Optimization}.
\newblock {\em CoRR\/} abs/1412.6980.

\bibitem[{Kiperwasser and Goldberg(2016)}]{TACL885}
Eliyahu Kiperwasser and Yoav Goldberg. 2016.
\newblock {Simple and Accurate Dependency Parsing Using Bidirectional LSTM
  Feature Representations}.
\newblock {\em Transactions of ACL\/} 4:313--327.

\bibitem[{K\"{u}bler et~al.(2009)K\"{u}bler, McDonald, and Nivre}]{Kubler2009}
Sandra K\"{u}bler, Ryan McDonald, and Joakim Nivre. 2009.
\newblock {\em {Dependency Parsing}\/}.
\newblock Synthesis Lectures on Human Language Technologies, Morgan \& cLaypool
  publishers.

\bibitem[{Kuncoro et~al.(2016)Kuncoro, Ballesteros, Kong, Dyer, and
  Smith}]{kuncoro-EtAl:2016:EMNLP2016}
Adhiguna Kuncoro, Miguel Ballesteros, Lingpeng Kong, Chris Dyer, and Noah~A.
  Smith. 2016.
\newblock {Distilling an Ensemble of Greedy Dependency Parsers into One MST
  Parser}.
\newblock In {\em Proceedings of EMNLP\/}. pages 1744--1753.

\bibitem[{Lapponi et~al.(2017)Lapponi, Oepen, and {\O}vrelid}]{Lap:Oep:Ovr:17}
Emanuele Lapponi, Stephan Oepen, and Lilja {\O}vrelid. 2017.
\newblock {EPE}~2017: {T}he {S}herlock negation resolution downstream
  application.
\newblock In {\em Proceedings of the EPE 2017 Shared Task\/}. pages 25--30.

\bibitem[{Lee et~al.(2011)Lee, Naradowsky, and
  Smith}]{Lee:2011:DMJ:2002472.2002584}
John Lee, Jason Naradowsky, and David~A. Smith. 2011.
\newblock {A Discriminative Model for Joint Morphological Disambiguation and
  Dependency Parsing}.
\newblock In {\em Proceedings of ACL-HLT\/}. pages 885--894.

\bibitem[{Li et~al.(2011)Li, Zhang, Che, Liu, Chen, and
  Li}]{Li:2011:JMC:2145432.2145557}
Zhenghua Li, Min Zhang, Wanxiang Che, Ting Liu, Wenliang Chen, and Haizhou Li.
  2011.
\newblock {Joint Models for Chinese POS Tagging and Dependency Parsing}.
\newblock In {\em Proceedings of EMNLP\/}. pages 1180--1191.

\bibitem[{Ma and Hovy(2017)}]{ma-hovy:2017:I17-1}
Xuezhe Ma and Eduard Hovy. 2017.
\newblock {Neural Probabilistic Model for Non-projective MST Parsing}.
\newblock In {\em Proceedings of IJCNLP\/}. pages 59--69.

\bibitem[{Marcus et~al.(1993)Marcus, Santorini, and
  Marcinkiewicz}]{Marcus93building}
Mitchell~P. Marcus, Beatrice Santorini, and Mary~Ann Marcinkiewicz. 1993.
\newblock {Building a Large Annotated Corpus of English: The Penn Treebank}.
\newblock {\em Computational Linguistics\/} 19(2):313--330.

\bibitem[{McDonald et~al.(2005)McDonald, Crammer, and
  Pereira}]{McDonald2005OLT}
Ryan McDonald, Koby Crammer, and Fernando Pereira. 2005.
\newblock Online large-margin training of dependency parsers.
\newblock In {\em Proceedings of ACL\/}. pages 91--98.

\bibitem[{McDonald and Pereira(2006)}]{E06-1011}
Ryan McDonald and Fernando Pereira. 2006.
\newblock {Online Learning of Approximate Dependency Parsing Algorithms}.
\newblock In {\em Proceedings of EACL\/}. pages 81--88.

\bibitem[{Neubig et~al.(2017)Neubig, Dyer, Goldberg, Matthews, Ammar,
  Anastasopoulos, Ballesteros, Chiang, Clothiaux, Cohn, Duh, Faruqui, Gan,
  Garrette, Ji, Kong, Kuncoro, Kumar, Malaviya, Michel, Oda, Richardson,
  Saphra, Swayamdipta, and Yin}]{dynet}
Graham Neubig, Chris Dyer, Yoav Goldberg, Austin Matthews, Waleed Ammar,
  Antonios Anastasopoulos, Miguel Ballesteros, David Chiang, Daniel Clothiaux,
  Trevor Cohn, Kevin Duh, Manaal Faruqui, Cynthia Gan, Dan Garrette, Yangfeng
  Ji, Lingpeng Kong, Adhiguna Kuncoro, Gaurav Kumar, Chaitanya Malaviya, Paul
  Michel, Yusuke Oda, Matthew Richardson, Naomi Saphra, Swabha Swayamdipta, and
  Pengcheng Yin. 2017.
\newblock {DyNet: The Dynamic Neural Network Toolkit}.
\newblock {\em arXiv preprint arXiv:1701.03980\/} .

\bibitem[{Nguyen et~al.(2017)Nguyen, Dras, and
  Johnson}]{nguyen-dras-johnson:2017:K17-3}
Dat~Quoc Nguyen, Mark Dras, and Mark Johnson. 2017.
\newblock {A Novel Neural Network Model for Joint POS Tagging and Graph-based
  Dependency Parsing}.
\newblock In {\em Proceedings of the CoNLL 2017 Shared Task\/}. pages 134--142.

\bibitem[{Nguyen and Verspoor(2018)}]{NguyenV2018}
Dat~Quoc Nguyen and Karin Verspoor. 2018.
\newblock {From POS tagging to dependency parsing for biomedical event
  extraction}.
\newblock {\em arXiv preprint arXiv:1808.03731\/} .

\bibitem[{Nivre(2003)}]{Nivre2003}
Joakim Nivre. 2003.
\newblock An efficient algorithm for projective dependency parsing.
\newblock In {\em Proceedings of IWPT\/}.

\bibitem[{Nivre et~al.(2018)Nivre, Abrams et~al.}]{11234/1-2837}
Joakim Nivre, Mitchell Abrams, et~al. 2018.
\newblock \href{http://hdl.handle.net/11234/1-2837}{{Universal Dependencies
  2.2}}.
\newblock
  \href{http://hdl.handle.net/11234/1-2837}{http://hdl.handle.net/11234/1-2837}.

\bibitem[{Nivre et~al.(2007{\natexlab{a}})Nivre, Hall, K\"{u}bler, McDonald,
  Nilsson, Riedel, and Yuret}]{Nivre07}
Joakim Nivre, Johan Hall, Sandra K\"{u}bler, Ryan McDonald, Jens Nilsson,
  Sebastian Riedel, and Deniz Yuret. 2007{\natexlab{a}}.
\newblock {The CoNLL 2007 Shared Task on Dependency Parsing}.
\newblock In {\em Proceedings of the CoNLL Shared Task Session of EMNLP-CoNLL
  2007\/}.

\bibitem[{Nivre et~al.(2007{\natexlab{b}})Nivre, Hall, Nilsson, Chanev,
  Eryigit, K\"{u}bler, Marinov, and Marsi}]{Nivre2007}
Joakim Nivre, Johan Hall, Jens Nilsson, Atanas Chanev, G\"{u}lsen Eryigit,
  Sandra K\"{u}bler, Svetoslav Marinov, and Erwin Marsi. 2007{\natexlab{b}}.
\newblock {MaltParser: A language-independent system for data-driven dependency
  parsing}.
\newblock {\em Natural Language Engineering\/} 13(2):95--135.

\bibitem[{Oepen et~al.(2017)Oepen, {\O}vrelid, Bj{\"o}rne, Johansson, Lapponi,
  Ginter, and Velldal}]{epe17}
Stephan Oepen, Lilja {\O}vrelid, Jari Bj{\"o}rne, Richard Johansson, Emanuele
  Lapponi, Filip Ginter, and Erik Velldal. 2017.
\newblock The 2017 {S}hared {T}ask on {E}xtrinsic {P}arser {E}valuation.
  {T}owards a reusable community infrastructure.
\newblock In {\em Proceedings of the EPE 2017 Shared Task\/}. pages 1--16.

\bibitem[{Pennington et~al.(2014)Pennington, Socher, and
  Manning}]{pennington-socher-manning:2014:EMNLP2014}
Jeffrey Pennington, Richard Socher, and Christopher Manning. 2014.
\newblock {Glove: Global Vectors for Word Representation}.
\newblock In {\em Proceedings of EMNLP\/}. pages 1532--1543.

\bibitem[{Plank et~al.(2016)Plank, S{\o}gaard, and Goldberg}]{plankP16}
Barbara Plank, Anders S{\o}gaard, and Yoav Goldberg. 2016.
\newblock {Multilingual Part-of-Speech Tagging with Bidirectional Long
  Short-Term Memory Models and Auxiliary Loss}.
\newblock In {\em Proceedings of ACL\/}. pages 412--418.

\bibitem[{Potthast et~al.(2014)Potthast, Gollub, Rangel, Rosso, Stamatatos, and
  Stein}]{tira}
Martin Potthast, Tim Gollub, Francisco Rangel, Paolo Rosso, Efstathios
  Stamatatos, and Benno Stein. 2014.
\newblock {Improving the Reproducibility of {PAN}'s Shared Tasks: Plagiarism
  Detection, Author Identification, and Author Profiling}.
\newblock In {\em Proceedings of CLEF Initiative\/}. pages 268--299.

\bibitem[{Reddy et~al.(2016)Reddy, T{\"a}ckstr{\"o}m, Collins, Kwiatkowski,
  Das, Steedman, and Lapata}]{Q16-1010}
Siva Reddy, Oscar T{\"a}ckstr{\"o}m, Michael Collins, Tom Kwiatkowski, Dipanjan
  Das, Mark Steedman, and Mirella Lapata. 2016.
\newblock {Transforming Dependency Structures to Logical Forms for Semantic
  Parsing}.
\newblock {\em Transactions of ACL\/} 4:127--141.

\bibitem[{Schuster et~al.(2017)Schuster, Clergerie, Candito, Sagot, Manning,
  and Seddah}]{stanfordparis2017}
Sebastian Schuster, Eric De~La Clergerie, Marie Candito, Benoit Sagot,
  Christopher~D. Manning, and Djam\'e Seddah. 2017.
\newblock {Paris and Stanford at EPE 2017: Downstream Evaluation of Graph-based
  Dependency Representations}.
\newblock In {\em Proceedings of the EPE 2017 Shared Task\/}. pages 47--59.

\bibitem[{Srivastava et~al.(2014)Srivastava, Hinton, Krizhevsky, Sutskever, and
  Salakhutdinov}]{JMLR:v15:srivastava14a}
Nitish Srivastava, Geoffrey Hinton, Alex Krizhevsky, Ilya Sutskever, and Ruslan
  Salakhutdinov. 2014.
\newblock {Dropout: A Simple Way to Prevent Neural Networks from Overfitting}.
\newblock {\em Journal of Machine Learning Research\/} 15:1929--1958.

\bibitem[{Straka and Strakov\'{a}(2017)}]{udpipe:2017}
Milan Straka and Jana Strakov\'{a}. 2017.
\newblock {Tokenizing, POS Tagging, Lemmatizing and Parsing UD 2.0 with
  UDPipe}.
\newblock In {\em Proceedings of the CoNLL 2017 Shared Task\/}. pages 88--99.

\bibitem[{Tateisi et~al.(2005)Tateisi, Yakushiji, Ohta, and Tsujii}]{I05-2038}
Yuka Tateisi, Akane Yakushiji, Tomoko Ohta, and Jun'ichi Tsujii. 2005.
\newblock {Syntax Annotation for the GENIA Corpus}.
\newblock In {\em Proceedings of IJCNLP: Companion Volume\/}. pages 220--225.

\bibitem[{Toutanova et~al.(2003)Toutanova, Klein, Manning, and
  Singer}]{N03-1033}
Kristina Toutanova, Dan Klein, Christopher~D. Manning, and Yoram Singer. 2003.
\newblock {Feature-Rich Part-of-Speech Tagging with a Cyclic Dependency
  Network}.
\newblock In {\em Proceedings of NAACL-HLT\/}.

\bibitem[{Weiss et~al.(2015)Weiss, Alberti, Collins, and
  Petrov}]{weiss-EtAl:2015:ACL-IJCNLP}
David Weiss, Chris Alberti, Michael Collins, and Slav Petrov. 2015.
\newblock Structured training for neural network transition-based parsing.
\newblock In {\em Proceedings of ACL-IJCNLP\/}. pages 323--333.

\bibitem[{Yamada and Matsumoto(2003)}]{Yamada2003}
Hiroyasu Yamada and Yuji Matsumoto. 2003.
\newblock {Statistical dependency analysis with support vector machines}.
\newblock In {\em Proceedings IWPT\/}. pages 195--206.

\bibitem[{Yang et~al.(2018)Yang, Zhang, Liu, Sun, Yu, and Fu}]{YangZLSYF18}
Liner Yang, Meishan Zhang, Yang Liu, Maosong Sun, Nan Yu, and Guohong Fu. 2018.
\newblock {Joint {POS} Tagging and Dependence Parsing With Transition-Based
  Neural Networks}.
\newblock {\em {IEEE/ACM} Trans. Audio, Speech {\&} Language Processing\/}
  26(8):1352--1358.

\bibitem[{Zeman et~al.(2018)Zeman, Ginter, Haji{\v{c}}, Nivre, Popel, and
  Straka}]{udst:overview}
Daniel Zeman, Filip Ginter, Jan Haji{\v{c}}, Joakim Nivre, Martin Popel, and
  Milan Straka. 2018.
\newblock {CoNLL 2018 Shared Task: Multilingual Parsing from Raw Text to
  Universal Dependencies}.
\newblock In {\em {Proceedings of the CoNLL 2018 Shared Task}\/}. page to
  appear.

\bibitem[{Zeman et~al.(2017)Zeman, Popel, Straka, Hajic, Nivre
  et~al.}]{K17-3001}
Daniel Zeman, Martin Popel, Milan Straka, Jan Hajic, Joakim Nivre, et~al. 2017.
\newblock {CoNLL 2017 Shared Task: Multilingual Parsing from Raw Text to
  Universal Dependencies}.
\newblock In {\em Proceedings of the CoNLL 2017 Shared Task\/}. pages 1--19.

\bibitem[{Zhang et~al.(2017)Zhang, Cheng, and
  Lapata}]{zhang-cheng-lapata:2017:EACLlong}
Xingxing Zhang, Jianpeng Cheng, and Mirella Lapata. 2017.
\newblock {Dependency Parsing as Head Selection}.
\newblock In {\em Proceedings of EACL\/}. pages 665--676.

\bibitem[{Zhang et~al.(2015)Zhang, Li, Barzilay, and
  Darwish}]{zhang-EtAl:2015:NAACL-HLT1}
Yuan Zhang, Chengtao Li, Regina Barzilay, and Kareem Darwish. 2015.
\newblock Randomized greedy inference for joint segmentation, pos tagging and
  dependency parsing.
\newblock In {\em Proceedings of NAACL-HLT\/}. pages 42--52.

\bibitem[{Zhang and Weiss(2016)}]{zhang-weiss:2016:P16-1}
Yuan Zhang and David Weiss. 2016.
\newblock {Stack-propagation: Improved Representation Learning for Syntax}.
\newblock In {\em Proceedings of ACL\/}. pages 1557--1566.

\bibitem[{Zhang and Nivre(2011)}]{zhang-nivre:2011:ACL-HLT2011}
Yue Zhang and Joakim Nivre. 2011.
\newblock Transition-based dependency parsing with rich non-local features.
\newblock In {\em Proceedings ACL-HLT\/}. pages 188--193.

\end{thebibliography}
\bibliographystyle{acl_natbib}

\end{document}